# A Novel Architecture for Computing Approximate Radon Transform

Mohammad Amin Khorsandi
Dept. of Elect. & Comp. Engineering
Isfahan Univ. of Technology,
Isfahan, Iran
ma.khorsandi@ec.iut.ac.ir

Nader Karimi
Dept. of Elect. & Comp. Engineering
Isfahan Univ. of Technology,
Isfahan, Iran
nader.karimi@cc.iut.ac.ir

Shadrokh Samavi
Dept. of Elect. & Comp. Engineering
Isfahan Univ. of Technology,
Isfahan, Iran
samavi96@cc.iut.ac.ir

*Abstract*—**Radon transform is a type of transform which is used in image processing to transfer the image into intercept-slope coordinate. Its diagonal properties made it appropriate for some applications which need processes in different degrees. Radon transform computation needs a lot of arithmetic operations which makes it a compute-intensive algorithm. In literature an approximate algorithm for computing Radon transform is introduces which reduces the complexity of computations. But this algorithm is complex and need arbitrary accesses to memory. In this paper we proposed an algorithm which accesses to memory sequentially. In the following an architecture is introduced which uses pipeline to reduce the time complexity of algorithm.**

I. INTRODUCTION

In the realm of digital image processing, Radon transform is a type of transform which represents images in (ρ,θ) coordination. Assuming parallel lines on an image, Radon transform is the integral of image data along those lines which ρ represents the intercept and θ represents the slope of the lines [1].

One of the most famous applications of Radon transform in medical sciences is in CT scanner devices [1]. The X-ray parallel beams projected from different angles to the object and received by an array of sensors. By acquiring all projection from different angles a 2D image is formed which is in intercept-slope coordination. But in fact in this application, the forward Radon transform is formed already and we need a backward transform to convert it to Cartesian coordination.

There are other applications which use the forward radon transform. In [2] the Radon transform is used to gain the direction of linear motion blur. This motion blur is caused by shaking the camera or moving objects in photography. A shape descriptor by Radon transform is represented in [3]. This shape descriptor is invariant to rotation, scale and translation. Authors in [4] use Radon transform in watermarking. In that work, instead of embedding the logo in main image, the logo is embedded in Radon transform of the image, which results better representation for edges. In [5] an improved type of Radon transform is used to extract the electricity transition lines in satellite images. Since Radon transform adds image pixels in different directions, it is appropriate to extract lines in images.

Computation of Radon transform is complex and compute-intensive. The required calculations are consist of add, multiply and trigonometric calculation. Several methods are presented for this issue. A famous software method is represented in [6]. This method achieves the best performance and is used by Matlab [7]. Assuming the image size of N×N and the P number of angles, this method has the time complexity of $O(P \times N^2)$ and the same number of calculation. A hardware implementation for Radon transform presented in [8]. In this work, p number of pipeline processing elements are used to reduce the time complexity of proposed method in [6] to $O(N^2)$. An approximate method is introduced in [9] which gains the time complexity of $O(N^2.\log N)$. Literature [10] represented an architecture for implementing the algorithm in [9]. Theoretically the minimum possible time complexity of this architecture is $O(\log N)$ but due to limitation in hardware and memory accesses, it's almost impossible. Another approach is used in [11]. Instead of calculating integral of image along lines with different angles, the integral is always calculated along straight lines but the input image will be rotated to that angle. It reduces the complexity of computing the Radon transform, but rotating the image has its own complexity. Since the rotation method is nearest neighbor, this would be an approximate method.

In this paper we introduce a novel architecture for computing approximate Radon transform which is based on pipeline and runs in parallel. In this architecture the data is loaded to pipeline row by row. In each stage of pipeline the row data is shifted, due to row number and related angle. After each stage of pipeline the shifted data of each row is collected and added by previous data of each angle. The time complexity of this method is $O(N)$.

The structure of this paper consists of four other sections. In Section II the basic notions of Radon transform is described. Section III consists of a modified approximate algorithm to reduce the complexity of algorithm in [9]. In Section IV, we represent an architecture for implementing the algorithm. In Section V the results of our algorithm is reported.

## II. RADON TRANSFORM

### A. background

Computation of Radon transform consists of adding data of image over straight lines. A series of parallel lines by the angle of θ is used to calculate integral over them. Another parameter is ρ which is intercept of each line. The equations for calculating Radon transform of continuous and discrete images (for image size of M×N) are shown in (1) and (2) respectively.

$$g(\rho,\theta) = \int_{-\infty}^{\infty}\int_{-\infty}^{\infty} f(x,y)\delta(x\cos\theta + y\sin\theta - \rho)d_x d_y \quad (1)$$

$$g(\rho,\theta) = \sum_{x=0}^{M-1}\sum_{y=0}^{N-1} f(x,y)\delta(x\cos\theta + y\sin\theta - \rho) \quad (2)$$

In this equation, $g(\rho_j, \theta_k)$ equal with the sum of image data over a line with the equation of $\rho_j = x\cos\theta_k + y\sin\theta_k$. Considering the discrete Radon transform, $g(\rho, \theta_k)$ is a projection in angle of θ, which forms a vector. This concept is illustrated in Fig. 1.

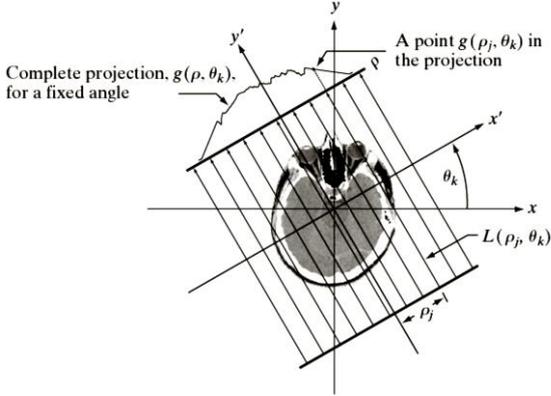

Fig. 1. The vector result of projection in the angle of $\theta_k$ [1].

By putting the vectors of different angles together, a 2D image is formed which is the result of Radon transform.

### B. Exact algorithm

During calculation of discrete Radon transform, it's impossible that projection lines pass exactly over the center of a pixel in all angles. Hence for an exact calculation, in each angle the distance between center of pixels and the projection line is calculated and gives a fraction number. This fraction number is used as percentage factor of that pixel to its related bin.

This notion is shown in Fig. 2. According to Fig. 2, the center of each pixel is projected then the space between each bin and projected location, defines the percentage effect of each pixel to the related bin. It should be noted that for increasing the accuracy of algorithm, each pixel is divided into 4 subpixels, then calculation is applied on subpixels [7].

For and N×N image with P number of projections, this algorithm has the time complexity of $O(P.N^2)$. the computation complexity of the algorithm is too high that number of required add and multiply operations are the same as time complexity.

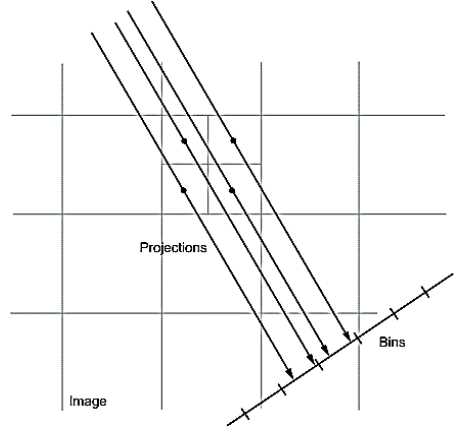

Fig. 2. The exact method for computing Radon transform [7].

### C. Approximate algorithm

Calculating the percentage effect of each pixel requires lots of multiply operation. To reduce the computations a type of approximate Radon transform is introduced in [9]. In this method the sum of pixels over a discrete line is considered as the projection bin. Six type of these lines in a 6×6 image are shown in Fig. 3.

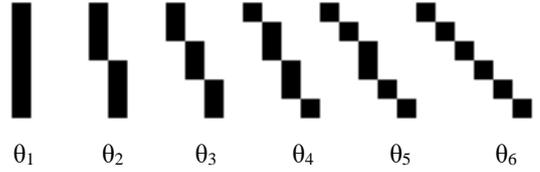

θ₁    θ₂    θ₃    θ₄    θ₅    θ₆

Fig. 3. Six type of discrete line from angle 0 to π/4 degrees for a 6×6 image.

Assuming the image size of N×N, the maximum possible angles between angles 0 and π/4 degrees are N angles. Clearly these lines are not covering a continuous line well, but as noted before, this is an approximate method.

The algorithm should add intensity of image pixels in different direction which these directions are defined by discrete lines in Fig. 3. For example; to compute 0 degree projection, all pixels in the same column must be added and forms a bin in 0 degree projection ($\theta_1$). The number of bins which contains a data is equal to N. For π/4 degree projection ($\theta_N$) each pixel must be added to the pixel on its bottom right, or in other word, the pixel located on the $\pi/4$ degree angle. The number of bins in this case is equal to $2N-1$. But for other projection ($\theta_2, \theta_3 ... \theta_{N-1}$) sometimes pixels must be add to their bottom pixel, and sometimes to the pixel on their bottom right. This concept for a 6×6 image is shown in Fig. 4.

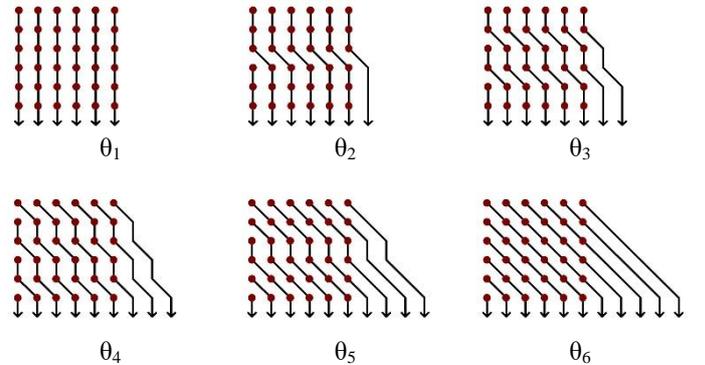

Fig. 4. Six type of projections in a 6×6 image. Each dot represents a pixel and each line shows the add order and each arrow shows the output bin.

## III. PROPOSED METHOD

Basically we use the idea of approximate Radon transform which is introduced in [9]. But we change the structure of the algorithm to make it possible to run in a pipeline.

As it shown in Fig. 4, two type of straight and diagonal add is required. To simplify the calculation we use a shift in each row and omit the directional add. In fact the input image is affine transformed and then the pixels in each column add together. This concept for a 6×6 image is shown in Fig. 5.

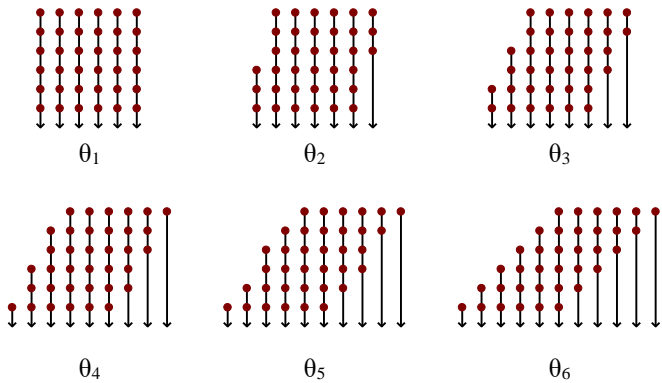

Fig. 5. Computing projection using affine transform of 6×6 image. Each dot represents a pixel and each line shows the add order and each arrow shows the output bin.

This algorithm is only for 0 to $\pi/4$ degree angles. For other angles a preprocessing stage is required. For 0 to $-\pi/4$ degree angles, the input data must be mirrored horizontally. For other angles, a transpose on the input data will prepare the data to be computed.

## IV. ARCHITECTURE

The architecture of proposed method consists of a pipeline with N stages. First, all rows are inserted to the pipeline row by row from top to bottom (each row data consists of the data of all pixels in same row). In each stage of pipeline there is a multiplexer. This multiplexer is responsible for passing input data directly or shifter by one pixel. For shifting data, all pixel data shifted to left by one pixel and a 0 8bit data (assuming 8bit for each pixel) is inserted to the rightmost pixel. This multiplexer is placed between registers of pipeline stages. A line equation calculator defines the control signal of multiplexer. This calculator has two inputs. An input is a constant number which determines the related angle, the other input is the row number. Considering these two inputs, the calculator decides to shift or not to shift row data. The shifted data must be added and saved for each angle. For this purpose, an adder and a register is used to calculate and save the data of each angle.

In the worst case the maximum shift on pixel of a row is equal to $2N-1$. Assuming the width of 8 for each pixel, the width of all adders, multiplexers and registers (except row number registers) are 8×($2N-1$) bits.

This architecture is shown in Fig. 6. In this architecture rows of image are inserted to the pipeline from top to bottom ($R_1, R_2 ... R_N$). At the same time the row numbers inserted to the pipeline in another group of registers. Depending on the row number and desired angle, they will be shifted or passed directly. For example, $R_1$ will not be shifted in any stages but row $R_N$ will be shifted in all stages (see Fig. 5). It take N clock pulsed for pipeline to be filled. After that in the N+1 clock pulse, the data of $\theta_1$ is computed completely. The data of other angles ($\theta_2, \theta_3 ... \theta_N$) will be prepared in next clock pulses respectively.

## V. VERIFICATION

To verify our design, we wrote a structural VHDL code in Xilinx ISE design suit and implemented our design. We used several standard images to evaluate our method. The result of our design and the result calculated by Matlab for a standard image of Lena is shown is Fig. 7.

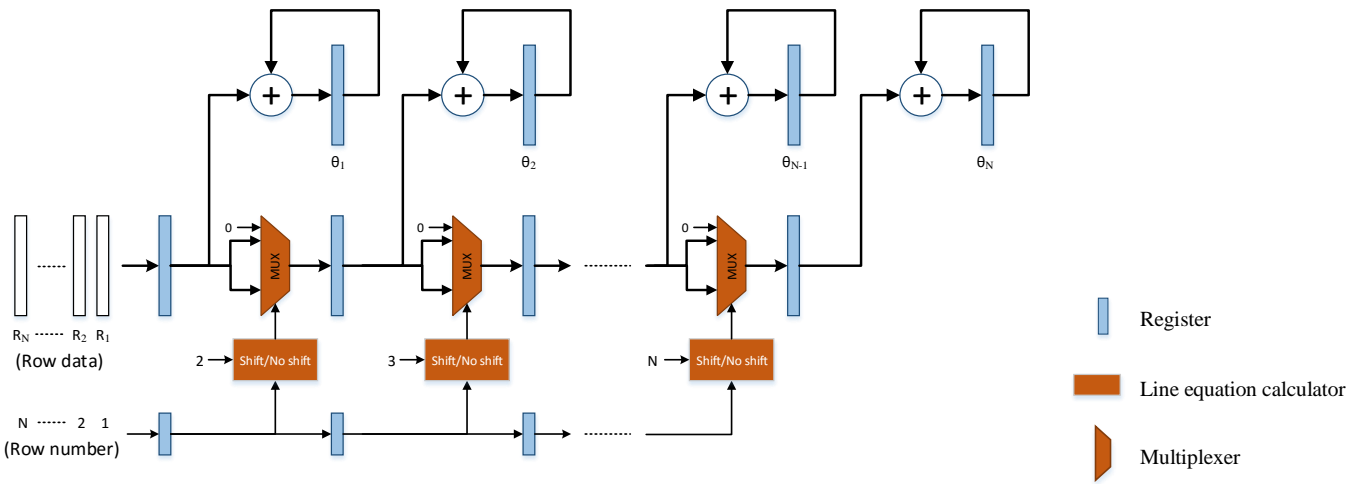

Fig. 6. Proposed architecture.

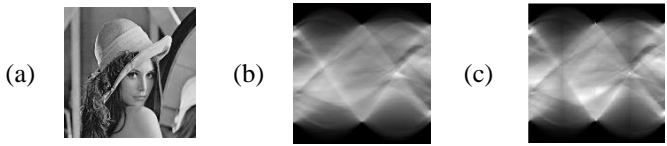

Fig. 7. (a) a standard image. (b) Radon transform of (a). (c) the result of our work

The main idea of approximate Radon transform in [9] has the time complexity of $O(N^2 \cdot \log N)$ and it claims that in case of using an $N^2$ array of processing elements, it is possible to gain the time complexity of $O(N)$. For this purpose, it is required to access all rows of image in memory arbitrarily. Literature [10] introduced and architecture for implementing [9]. The pattern of memory access for each row in this architecture is shown in Fig. 8.

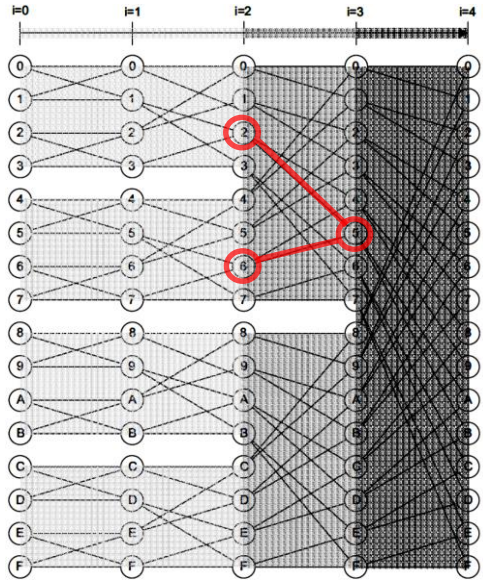

Fig. 8. Memory access pattern of proposed architecture in [10] for a 16×16 image. Each line represents a transfer in memory data.

Each line in Fig. 8 represents a transfer in memory data. As an example, we defined a transfer in this figure by red color. In this transfer the data in the row 6 must be shifted then added to data in row 2 and the result will be saved in row 5. Clearly this transfers makes the architecture complex and this complexity increases in large images. Due to limitations in memory accesses, gaining the time complexity of $O(\log N)$ for large images is almost impossible. Furthermore architecture in [10] faces a few problems such as address aliasing that causes read-after-write (RAW) error. Also in some iterations a part of hardware is idle and other part of hardware must be active for more iterations. But in our architecture the row data accessed sequentially in a pipeline that gives the time complexity of $O(N)$. This architecture is notably simple and due to using pipeline, all part of architecture is always active. It is also available to use a continuous stream of data for some cases such as video frames.

## VI. CONCLUSION

We designed an architecture for implementing approximate Radon transform which uses a pipeline to achieve the time complexity of $O(N)$. For computing projection data, we shift each row of image in case of necessity, then add it to other rows. Comparing to architecture in [10] our design is much simple. Our design needs to access memory data sequentially whereas [10] accesses memory data arbitrarily which makes this architecture complex.